\documentclass{article} %

\usepackage[final]{gpsmdms_2022}

\usepackage[american]{babel}
\usepackage{natbib}
\usepackage{microtype}
\usepackage{graphicx}
\usepackage{caption}
\usepackage{subcaption}
\usepackage{booktabs}
\usepackage{hyperref}

\usepackage{amsmath}
\usepackage{amssymb}
\usepackage{mathtools}
\usepackage{amsthm}
\usepackage{multicol}
\usepackage[capitalize,noabbrev]{cleveref}
\usepackage{xfrac}
\usepackage{algorithm}
\usepackage{algpseudocode}
\usepackage{multicol}
\usepackage{float}
\usepackage{pdfpages}

\usepackage{tikz}
\usetikzlibrary{bayesnet}

\title{Ice Core Dating using Probabilistic Programming}

\author{%
    Aditya Ravuri \\
    University of Cambridge \\
    \texttt{ar847@cam.ac.uk}
    \And
    Tom R. Andersson \\
    British Antarctic Survey \\
    \texttt{tomand@bas.ac.uk}
    \And
    Ieva Kazlauskaite \\
    University of Cambridge \\
    British Antarctic Survey \\
    \texttt{ik394@cam.ac.uk}
    \And
    Will Tebbutt \\
    University of Cambridge \\
    \texttt{wct23@cam.ac.uk}
    \And
    Richard E. Turner \\
    University of Cambridge \\
    \texttt{ret26@cam.ac.uk}
    \And
    J. Scott Hosking \\
    British Antarctic Survey \\
    \texttt{jask@bas.ac.uk}
    \And
    Neil D. Lawrence \\
    University of Cambridge \\
    \texttt{ndl21@cam.ac.uk}
    \And
    Markus Kaiser \\
    University of Cambridge \\
    British Antarctic Survey \\
    \texttt{mk2092@cam.ac.uk}
}

\begin{document}
\maketitle

\begin{abstract}
Ice cores record crucial information about past climate.
However, before ice core data can have scientific value, the chronology must be inferred by estimating the age as a function of depth.
%
Under certain conditions, chemicals locked in the ice display quasi-periodic cycles that delineate annual layers.
%
%
Manually counting these noisy seasonal patterns to infer the chronology can be an imperfect and time-consuming process, and does not capture uncertainty in a principled fashion.
In addition, several ice cores may be collected from a region, introducing an aspect of spatial correlation between them.
We present an exploration of the use of probabilistic models for automatic dating of ice cores, using probabilistic programming to showcase its use for prototyping, automatic inference and maintainability, and demonstrate common failure modes of these tools.
\end{abstract}

\section{Introduction}
\label{section:intro}

\begin{figure}[t]
    \centering
    \includegraphics[width=0.75\textwidth]{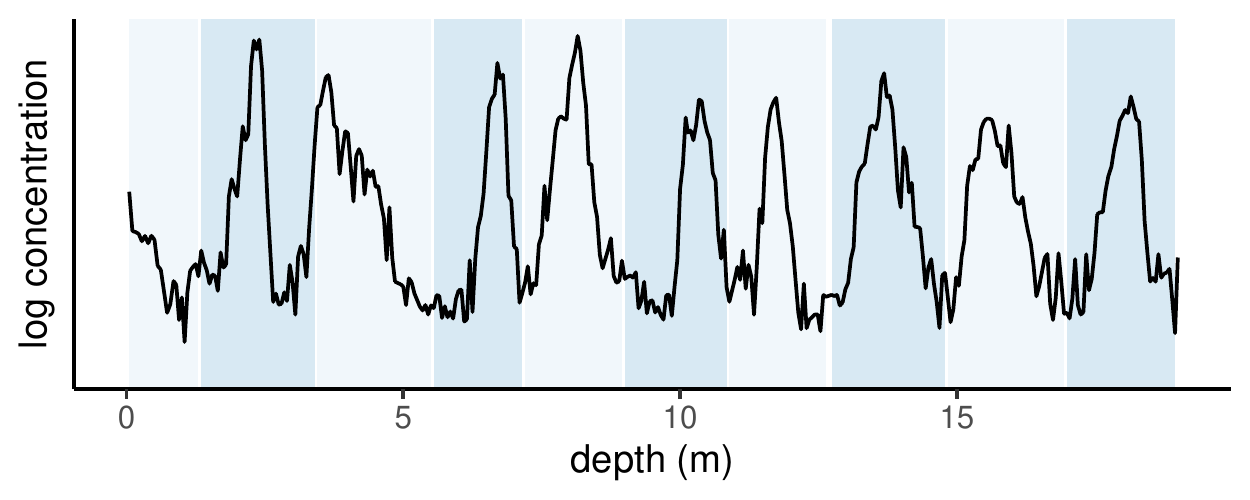}
    \caption{
        Section of the data, showing annual seasonality in MSA (a proxy exhibiting a strong seasonal pattern) plotted against depth. Manually counted annual layers are overlaid as shaded bars.
        \label{fig:data}
    }
\end{figure}

Chemicals in the atmosphere are deposited onto ice sheets through precipitation, with further deposition burying and eventually compacting the snow into solid ice, recording the chemical composition of the atmosphere. These chemicals provide evidence for the climate of the past and are known as \emph{proxy variables}.
Annual cycles can be present in the data if 1) the abundance of that proxy varies seasonally, 2) the precipitation rate at the ice core site is large enough, and 3) the depth is not so great that the annual layers have been excessively compressed. Given these conditions, annual layer thickness is dictated by the amount of annual precipitation (a random component) and compression of the ice with increasing depth (a systematic component). A section of the data that we used, the Jurassic ice core from \cite{bas-data}, is shown in \cref{fig:data}.

Constructing an ice core's timescale manually through layer counting can be an arduous process, lasting from days to years of person-time (\citealt{winstrup_hidden_2016}). Manual counting also has the disadvantage of poorly-quantified uncertainty in the depth to time mapping, based on heuristics from expert disagreement or uncertain layers. Probabilistic inference leads to a more principled treatment of such uncertainties.
A probabilistic approach also enables prior knowledge to be incorporated during ice core timescale inference. For example, some observations for time at certain depths may be available due to known volcanic events depositing a layer of ash in the ice (these depth-time observations are known as tie-points).


Our contribution is to present a case study in developing models for the ice core dating problem using a Probabilistic Programming Language (PPL), mainly Stan \citep{stan} alongside torchsde \citep{torchsde}, and TensorFlow Probability \citep{tfp}.
Our work highlights such tools to practitioners at the intersection of probabilistic machine learning and climate science.
We aim to show the promise of PPLs in how they can enable the composition of assumptions, easy experimentation, model extension and maintainability.
A core promise of widely used PPLs is that they automate and abstract away inference details, thus ensuring that models are written at the right level of abstraction without the need to write and maintain complex inference algorithms.
We also show current limitations of these tools, such as inference only being computationally feasible over a limited set of models that may be suitable for a task, and inference methods being limited in practice depending on the specific functional requirements set out for each PPL.

\section{Problem Setup}
\label{section:setup}

The proxy variables are measured in ice cores along a depth dimension. We represent the depth series as a set of known random variables $\boldsymbol{\delta} \coloneqq \{\delta_i\}_{i=1}^n, \delta_i \in \mathbb R^+$ where $n$ is the number of sampled depths at which proxy readings are available. The depth series $\boldsymbol{\delta}$ is mapped to latent time (age) values associated with each observation, represented by the stochastic process $\mathbf{t} \coloneqq \{t_{\delta_i}\}_{i=1}^n$ indexed by the depth series.
Proxy measurements are denoted as $\mathbf{s} \coloneqq \{s_{\delta_i}\}_{i=1}^n$, with $s_{\delta_i} \in \mathbb{R}$.
The \emph{ice core dating} process can be stated as an inference problem for time conditioned on the proxy and depth data, $\mathbf{t} | \mathbf{s}, \boldsymbol{\delta}$.
Within this work, we assume that proxies depend only on the latent time process, the depth to time mapping is monotonic, and that there is only one proxy available with a clear seasonal signal.
This gives rise to the class of models shown in Fig.~\ref{fig:models} (a), in the Appendix.

\section{Methodology}
The setup of the problem gives rise to a class of models that vary in their assumptions and complexity from discrete-index HMMs to continuous-index HMMs to SDE-based models.
Some of these models, particularly HMMs, were previously studied by~\cite{mw_thesis, winstrup_hidden_2016} who explored the the effect of batching, the usage of other observation models and extensions to hidden semi-Markov models (for allowing for priors to be set over lengths of year boundaries) and inference therein. 
\subsection{A Hidden Markov Model}
\label{subsec:simple-hmm}

\paragraph{Model} Under the assumption that the depth-sampling is uniform, and that the latent time process has a discrete domain, we can model the latent time process as a Markov chain. If in addition, the proxies are conditionally independent given the time periods they correspond to, the framework in \cref{fig:models} (a) reduces to a Hidden Markov Model shown in \cref{fig:models} (b). We assume that $\boldsymbol{t}$ can occupy states $\forall i: t_{\delta_i} \in \{\sfrac{k}{n_s}\}_{k=1}^{m \cdot n_s}$, where $n_s$ is the number of states within each yearly cycle and $m$ is an arbitrarily large number of years. The transition matrix
\begin{equation}
\label{eqn:transmat}
\mathbb{P}(t_{\delta_i}|t_{\delta_{i-1}}) =  \begin{bmatrix}
p_{1/n_s} & 1 - p_{1/n_s} & 0 & \cdots & 0 \\
0 & p_{2/n_s} & 1 - p_{2/n_s} & \cdots & 0 \\
0 & 0 & 0 & 0 & 1 \\
\end{bmatrix}
\end{equation}
is bidiagonal, with $p_i$ denoting the probability that the chain stays in state $i$ given that it was in the same state at the last depth measurement. In other words, as depth increases, the time process either advances or stays the same.
This enforces monotonicity of the latent time sequence w.r.t.\ depth and also allows one to track what year an observation corresponds to (given by the floor of the state, $\lfloor i \rfloor$). The parameters for each state within a year are repeated across years to constrain the model (i.e. $p_{c + k/n_s} = p_{k/n_s}$ for $c \in \mathbb N$). The following observation model is used as part of the HMM,
$$ \forall i: s_{\delta_i} | t_{\delta_i} \sim \mathcal{N}(a\cos(2\pi t_{\delta_i}) + b, \sigma^2). $$
Note that, as the transition matrix parameters are not constant within each annual layer, the model's annual layer shapes can be a warping of the mean cosine function in the observation model, enabling some flexibility in modelling the real proxy cycle shapes.

\paragraph{Inference} Using the Stan language, we perform maximum likelihood inference for the parameters given data with time marginalised ($a, b, \sigma, \{p\}_{j}|\mathbf{s}$), with the posterior over times $\mathbf{t} | \mathbf{s}, a, b, \sigma, \{p\}_{j}$ estimated using the forward-backward algorithm in Stan. The probabilistic program is shown in \cref{app:stan-simple-hmm}. Two problems are encountered with standard tooling:
\begin{itemize}
    \item The runtimes increase quadratically w.r.t.\ the state space (due to the forward algorithm, which computes the log likelihood with hidden states marginalised). We remedy this by rewriting the forward algorithm exploiting the sparsity of our transition matrix, providing a direct replacement of the native Stan function (shown in \cref{app:stan-simple-hmm}). In a PPL, users can typically implement their own efficient functions for such likelihood calculations.
    \item The model is misspecified as the data is expected to be non-stationary (due to compression of the ice core and temporal variation in precipitation). This can pragmatically be remedied by processing the data in batches, allowing the parameters such as $a$ and $b$ to change between different sections of the ice core.
\end{itemize}
Inference using this implementation takes 2.5 minutes using a single-thread run, without GPU utilization, for the entire ice core (about 2.5k observations), without requiring any special initialization strategies.

\subsection{An extension allowing for tie-point specification}
\label{subsec:tie-points}

To allow for tie-points to be integrated, which constrain the depth-to-time mapping, 
the observation model is extended to account for non-stationarity in the signal, because accounting for tie-points would require processing data in batches large enough to cover the tie-points. This is done by changing the observation model to allow for parameters $a, b$ to change with each data point along the ice core,
\begin{align*}
    \forall i: s_{\delta_i} | t_{\delta_i} &\sim \mathcal{N}(a_i\cos(2\pi t_{\delta_i}) + b_i, \sigma^2),
\end{align*}
with a prior (as part of a hierarchical model) placed over $a_i, b_i$ (thus, the change is ``slow''/constrained). A similar hierarchical treatment is given to parameters of the transition matrix, using the prior
$$ \forall j \in 1/n_s, ..., m: p_{j} \sim \text{Beta}(\alpha_{j*n_s \text{ mod } n_s}, \beta_{j*n_s \text{ mod } n_s}).$$
This allows the transition probabilities to change over years (and hence depths). The hierarchical distribution of states at the same point in any yearly cycle however share the same parameters, providing some constraint.

Having extended the model to different observation models per data-point, we specify volcanic tie-points using an alternative observation model using the state directly rather than proxy information.
We use a categorical distribution ensuring that the time states must reach the volcanic tie-points,
$$ p(s'_{\delta_{\text{tie}}} | t_{\delta_{\text{tie}}}) = \begin{cases} 1/n_s & \text{if }  \lfloor t_{\delta_{\text{tie}}} \rfloor = t_{\text{tie}} \\
0 & \text{otherwise} \end{cases}, $$
to enforce that the volcanic ash observed at depth $\delta_{\text{tie}}$ corresponds to a known year of the eruption $t_{\text{tie}}$.

\begin{figure*}
\centering
\begin{minipage}[b]{.45\textwidth}
    \centering
    \includegraphics[width=1\textwidth]{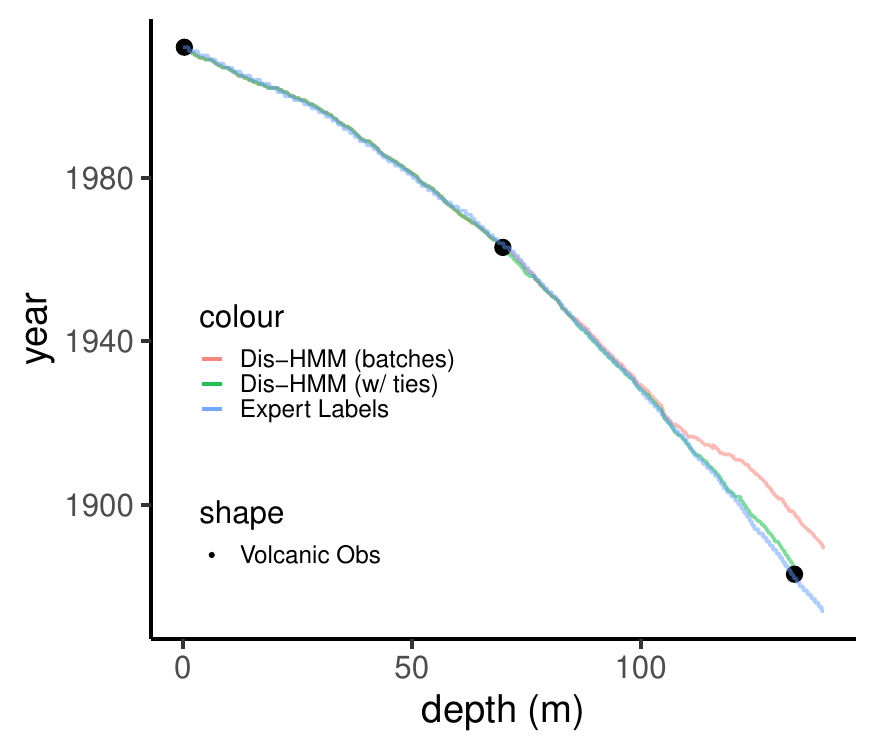}
    \caption{A comparison between the inference results, i.e. sample paths of $\mathbf{t} | \mathbf{s}$, obtained using Stan. These correspond to models presented in sections \ref{subsec:simple-hmm} and \ref{subsec:tie-points}, with expert labels overlaid.}\label{fig:comparisons}
\end{minipage}
\qquad
\begin{minipage}[b]{.45\textwidth}
    \centering
    \includegraphics[width=0.9\textwidth]{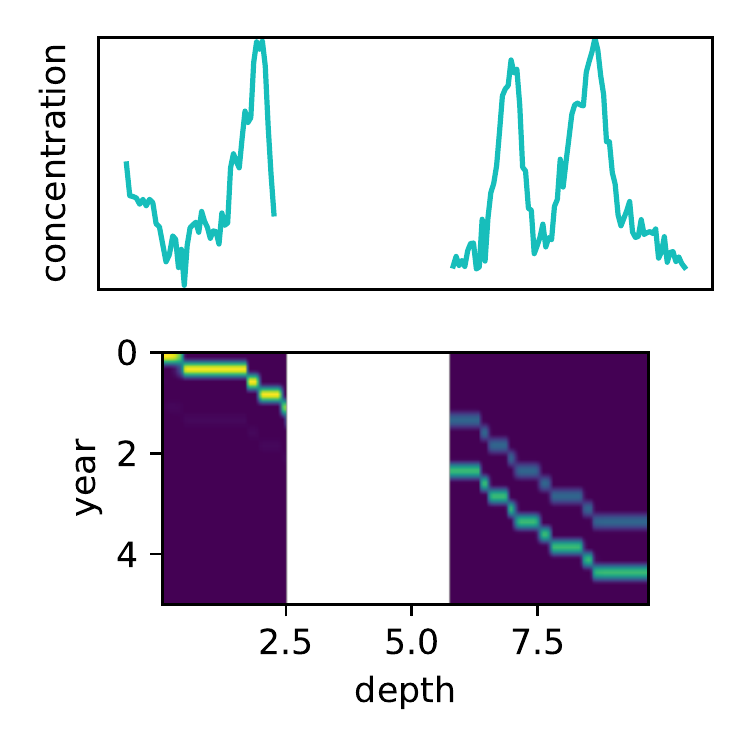}
    \caption{On top, a short data series with a missing section. At the bottom, posterior paths of time inferred using the data above - the missing section induces multimodality in the depth to time posterior.}\label{fig:cts-hmm}
\end{minipage}
\end{figure*}

Maximum likelihood inference for parameters such as $a_i, b_i$ in such models produces subpar results, due to posterior modes of such unidentifiable models lying outside their typical sets, raising a need to integrate out the parameters. As MCMC is computationally infeasible due to the number of parameters and data, we use mean-field variational inference (a seamless change in the Stan API). This mode of inference greatly increases runtimes w.r.t.~the model in \cref{subsec:simple-hmm} however, to about a few hours due to the increased number of parameters.
Results of inference in the two models presented thus far is shown in \cref{fig:comparisons}.

\subsection{An extension to continuous-index Hidden Markov Models}
\label{subsec:cts-hmm}

Depth measurements may not be regularly spaced, due to missing sections of an ice core as a result of the extraction process, or due to the sampling frequency changing over its depth.
In this case, modelling the latent time process as a continuous-index Markov chain leads to the times observed at irregularly spaced points being described by an index-(depth) inhomogeneous Markov chain.
Here, the transition matrix associated with a step from $\delta_{i-1}$ and $\delta_{i}$ can be computed via the transition rate matrix $\mathbf{Q}$ with the same sparsity structure as Equation \ref{eqn:transmat},
\begin{equation}
    \mathbb{P}(t_{\delta_i} | t_{\delta_{i-1}}) = \exp_{\text{matrix}}((\delta_i - \delta_{i-1}) \mathbf{Q}).
    \label{eqn:cts-trans}
\end{equation}
Such models are termed \emph{continuous-time HMMs} \citep{cts-time-hmm} (although we use the terminology continuous-index to avoid confusion as time is not the index in our application). They allow a better representation of posterior uncertainty arising from missing observations, as seen in Fig.~\ref{fig:cts-hmm}. Inference was performed on small datasets by using tensorflow-probability for computing log likelihoods of HMMs in a differentiable manner. High-level pseudocode for MLE/VI when using such tools is shown in \cref{app:other-models}. Inference involving larger datasets would involve bespoke inference code due to a lack of functionality around efficient computation of the forward algorithm involving matrix exponentials within the PPLs considered.

\subsection{An exploration of Stochastic Differential Equations for ice core dating}

Assuming instead that we are working with a continuous index and a continuous state space for $\mathbf{t}$, the class of models under consideration naturally extends to state space/ SDE models. Given an index~$\delta$, a prior over a stochastic process $t_{\delta}$ can be formulated as,
$$ \begin{bmatrix} d\mathbf{z}_\delta \\ dt_\delta \end{bmatrix} = \begin{bmatrix} \mu(\mathbf{z}_\delta, \delta) \\ -\exp(\mu'(\mathbf{z}_\delta, t_\delta, \delta)) \end{bmatrix} d\delta + \begin{bmatrix} \Sigma \\ \epsilon \end{bmatrix} d\mathbf{W}_\delta, $$
with $\epsilon \rightarrow 0$ enforcing monotonicity of sample paths and where $\mathbf{z}_{\delta}$ is a latent process that, for example, can have a GP prior represented as an SDE \citep{sarkkasolin}. Such a prior is similar to the one used in \cite{mon-gp-flow}. Inference in this class of models can be performed by specifying a variational SDE using the same diffusion but with a different drift \citep{torchsde}. As in the case above, due to customised variational inference not being supported in Stan, we use functionality from torchsde (differentiable solvers) that enable computation of the objective.

Inference in this class of models was particularly difficult, mirroring the findings of GPCore \citep{gpcore}, where a GP prior is placed on $\mathbf{t}|\boldsymbol{\delta}$, and where maximum likelihood inference is performed for $\mathbf{t}|\mathbf{s},\mathbf{d}$, in a constrained manner to ensure monotonicity in the depth-time sample paths. The inference is very dependent on good initialisation; using existing estimates of the chronology in the case of GPCore, and using inverse Lomb-Scargle spectrograms for $\mathbf{z}_{\delta}$ in our SDE models, following ideas from \cite{tf-analysis}. However, we had greater difficulties due to local minima than in GPCore. More discussion on this topic can be found in \cref{app:sde-results}.


\section{Conclusion}
In this paper we presented a taxonomy of models for ice core dating, showing how simplifying assumptions lead naturally to HMMs and how these models, and corresponding probabilistic programs can be extended. We exemplified the shortcomings of PPLs as the models are extended towards SDEs and discussed difficulties with inference in such models. Future work would involve extension of the models presented to utilize multiple proxies, account for spatial correlation between ice cores, and an exploration to determine how to aid inference in latent SDE models for such modelling tasks.


\subsection*{Data and Code}

Code to reproduce key results in this paper can be found at \href{https://github.com/infprobscix/icecores}{https://github.com/infprobscix/icecores}.

\subsection*{Acknowledgements}

AR is supported by a studentship from the Accelerate Programme for Scientific Discovery. IK was funded by a Biometrika Fellowship awarded by the Biometrika Trust. The authors would also like to thank Liz Thomas and Dieter Tetzner from the British Antarctic Survey for providing the datasets we used, for their insight and helpful discussions on ice core dating task.

\newpage
\bibliography{refs}

\begin{thebibliography}{14}
\providecommand{\natexlab}[1]{#1}
\providecommand{\url}[1]{\texttt{#1}}
\expandafter\ifx\csname urlstyle\endcsname\relax
  \providecommand{\doi}[1]{doi: #1}\else
  \providecommand{\doi}{doi: \begingroup \urlstyle{rm}\Url}\fi

\bibitem[Andersson(2019)]{gpcore}
Tom Andersson.
\newblock {GPCore}: A {Gaussian} process approach for inferring ice core
  chronologies.
\newblock Master's thesis, University of Cambridge, 2019.

\bibitem[Barber(2012)]{barber}
David Barber.
\newblock \emph{{Bayesian Reasoning and Machine Learning}}.
\newblock {Cambridge University Press}, 2012.

\bibitem[Dillon et~al.(2017)Dillon, Langmore, Tran, Brevdo, Vasudevan, Moore,
  Patton, Alemi, Hoffman, and Saurous]{tfp}
Joshua~V. Dillon, Ian Langmore, Dustin Tran, Eugene Brevdo, Srinivas Vasudevan,
  Dave Moore, Brian Patton, Alex Alemi, Matt Hoffman, and Rif~A. Saurous.
\newblock Tensorflow distributions, 2017.

\bibitem[Emanuelsson et~al.(2022)Emanuelsson, Thomas, Tetzner, Humby, and
  Vladimirova]{bas-data}
B.~Daniel Emanuelsson, Elizabeth~R. Thomas, Dieter~R. Tetzner, Jack~D. Humby,
  and Diana~O. Vladimirova.
\newblock Ice core chronologies from the antarctic peninsula: The palmer,
  jurassic, and rendezvous age-scales.
\newblock \emph{Geosciences}, 12\penalty0 (2), 2022.

\bibitem[Gay et~al.(2014)Gay, De~Angelis, and Lacoume]{tf-analysis}
M.~Gay, M.~De~Angelis, and J.-L. Lacoume.
\newblock Dating a tropical ice core by time–frequency analysis of ion
  concentration depth profiles.
\newblock \emph{Climate of the Past}, 10\penalty0 (5), 2014.

\bibitem[Hensman et~al.(2016)Hensman, Durrande, and Solin]{rff}
James Hensman, Nicolas Durrande, and Arno Solin.
\newblock Variational fourier features for gaussian processes, 2016.

\bibitem[Li et~al.(2020)Li, Wong, Chen, and Duvenaud]{torchsde}
Xuechen Li, Ting-Kam~Leonard Wong, Ricky T.~Q. Chen, and David Duvenaud.
\newblock Scalable gradients for stochastic differential equations, 2020.

\bibitem[Liu et~al.(2015)Liu, Li, Li, Song, and Rehg]{cts-time-hmm}
Yu-Ying Liu, Shuang Li, Fuxin Li, Le~Song, and James~M. Rehg.
\newblock Efficient learning of continuous-time hidden markov models for
  disease progression.
\newblock \emph{Advances in Neural Information Processing Systems},
  28:\penalty0 3599--3607, 2015.

\bibitem[Murphy(2012)]{pml}
Kevin~P Murphy.
\newblock \emph{Machine learning{:} a probabilistic perspective}.
\newblock 2012.

\bibitem[{Stan Development Team}(2022)]{stan}
{Stan Development Team}.
\newblock Stan modeling language users guide and reference manual 2.29, 2022.
\newblock URL \url{https://mc-stan.org}.

\bibitem[Särkkä and Solin(2019)]{sarkkasolin}
Simo Särkkä and Arno Solin.
\newblock \emph{Applied Stochastic Differential Equations}.
\newblock Institute of Mathematical Statistics Textbooks. 2019.

\bibitem[Ustyuzhaninov et~al.(2020)Ustyuzhaninov, Kazlauskaite, Ek, and
  Campbell]{mon-gp-flow}
Ivan Ustyuzhaninov, Ieva Kazlauskaite, Carl~Henrik Ek, and Neill Campbell.
\newblock Monotonic {G}aussian process flows.
\newblock In \emph{International Conference on Artificial Intelligence and
  Statistics}, pages 3057--3067. PMLR, 2020.

\bibitem[Winstrup(2011)]{mw_thesis}
Mai Winstrup.
\newblock \emph{An Automated Method for Annual Layer Counting in Ice Cores}.
\newblock PhD thesis, University of Copenhagen, 2011.

\bibitem[Winstrup(2016)]{winstrup_hidden_2016}
Mai Winstrup.
\newblock A hidden markov model approach to infer timescales for
  high-resolution climate archives.
\newblock In \emph{Proceedings of the Thirtieth {AAAI} Conference on Artificial
  Intelligence}, {AAAI}'16, pages 4053--4060. {AAAI} Press, 2016.

\end{thebibliography}

\clearpage
\appendix

\section{Appendix: Supplementary Figures}
\label{app:supp-figs}

\begin{figure}[h]
\centering
    \begin{tikzpicture}
    \node[obs] (d) {$\boldsymbol{\delta}$};
    \node[latent, right=of d] (t) {$\mathbf{t}$};
    \node[obs, right=of t] (s) {$\mathbf{s}$};
    
    \edge{d}{t}; 
    \edge{t}{s}; 
    
    \end{tikzpicture}
    \qquad \qquad \qquad \vline \qquad \qquad \qquad
    \begin{tikzpicture}
      \node[obs](y1) {$s_{\delta_1}$};
      \node[obs, right=of y1] (y2) {$s_{\delta_2}$};
      \node[obs, right=of y2] (y3) {$...$};
      \node[latent, above=of y1](t1) {$t_{\delta_1}$};
      \node[latent, above=of y2](t2) {$t_{\delta_2}$};
      \node[latent, above=of y3](t3) {$...$};
    
      \edge{t1}{y1};
      \edge{t1}{t2};
      \edge{t2}{t3};
      \edge{t2}{y2};
      \edge{t3}{y3};
    \end{tikzpicture}
\caption{
    (a.) On the left, a graph that summarises the class of models under consideration.
    (b.) On the right, a graph representing a hidden Markov model obtained when assumptions of the form in \cref{subsec:simple-hmm} are made.
    \label{fig:models}
}
\end{figure}
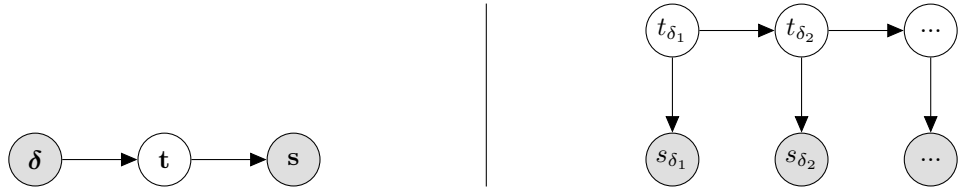

\section{Appendix: Discrete-index HMMs in Stan}
\label{app:stan-simple-hmm}

Probabilistic programs in Stan are composed as blocks corresponding to,
\begin{itemize}
    \item data and transformed data: where users can input / specify random variables for which observations have been made and other fixed values,
    \item parameters and transformed parameters: where users specify random variables on which inference is performed,
    \item model: where users specify distributional assumptions on and between the random variables defined,
    \item generated quantities: where downstream analysis can be computed as a function of posterior draws of the parameters conditioned on data.
\end{itemize}
Extensive documentation on example models in the Stan language, on functions and the language can be found \href{https://mc-stan.org/users/documentation/}{here}

We define the data blocks in Stan as follows, for all use cases.

\includegraphics{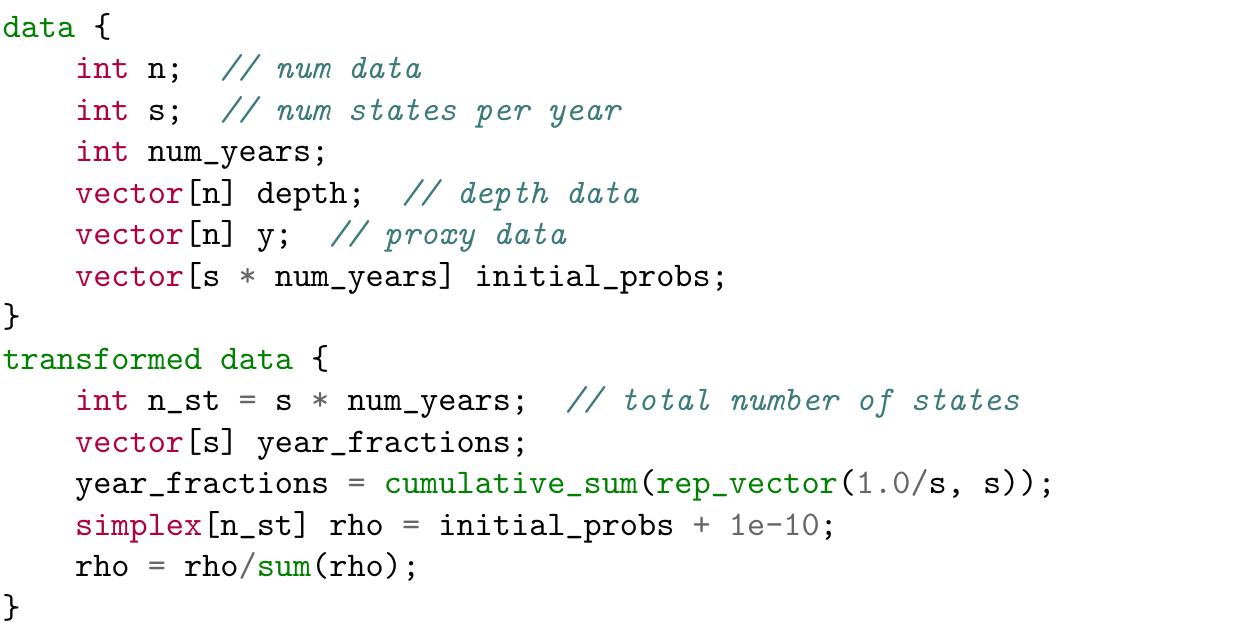}

A simple HMM in Stan, corresponding to the model defined in \cref{subsec:simple-hmm}, would consist of parameters,

\includegraphics{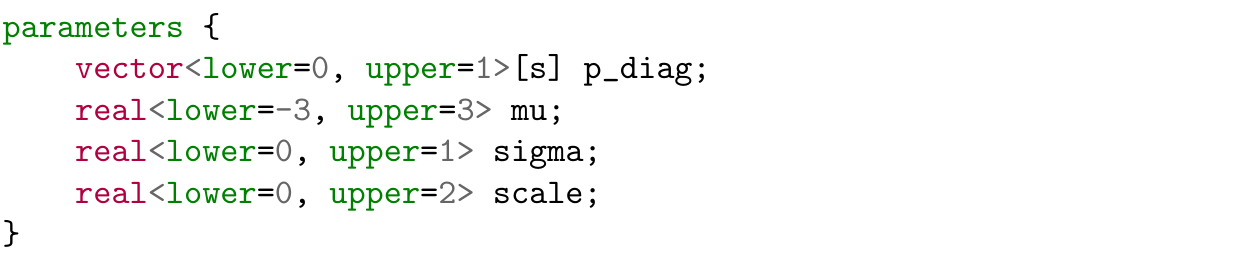}

on which inference will occur. Following the \href{https://mc-stan.org/docs/functions-reference/hmm-stan-functions.html}{stan syntax} on how HMMs are defined, we define \texttt{log\_omega} corresponding to our observation model, and other variables as follows,

\includegraphics{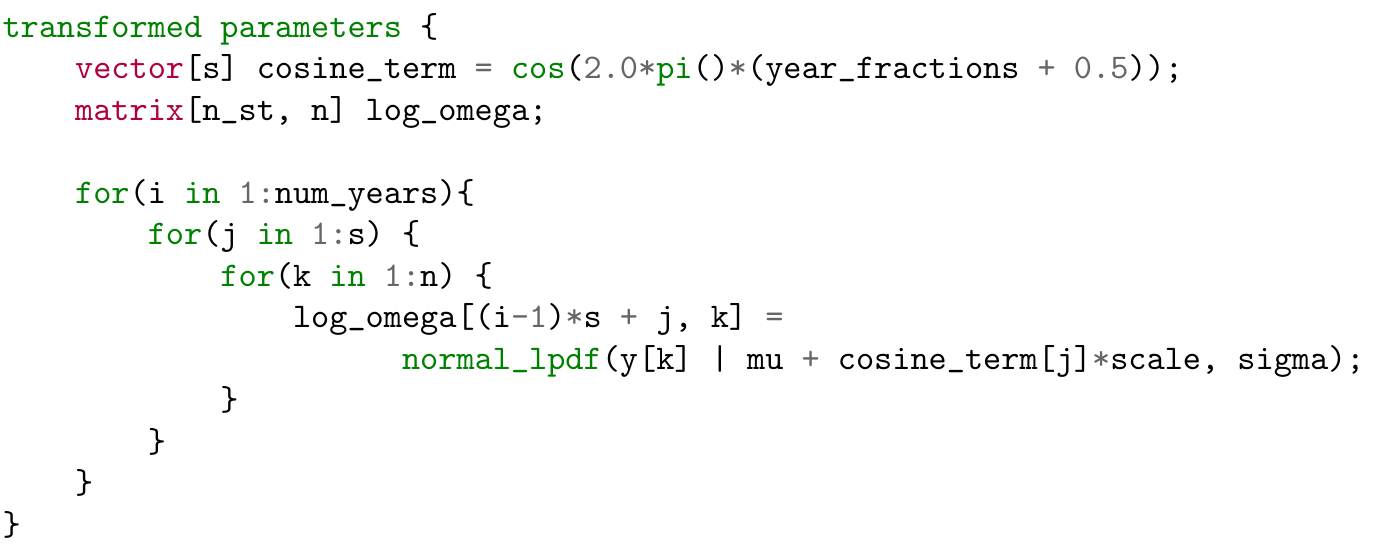}

Note that we create these variables in the transformed parameters section instead of the model section (which would be more Stan like, as it would be easier to read the observation model in the model section) to be able to access these variables in the generated quantities block without having to redefine them. It's an unusual case syntactical case as that the parameters governing the likelihood of the HMM with hidden states marginalised is a function of the observation model likelihood. Nevertheless, the observation model is specified in the line,

\includegraphics{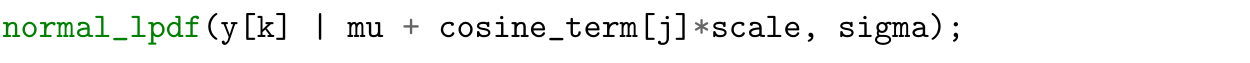}

The model then simply specifies the log posterior, which is a sum of priors over our parameters (implicitly assumed to be uniform as they're not specified) and the likelihood of
$$\mathbf{s} | \mu, \sigma, \text{scale and transition matrix diagonal}$$
which is specified as,

\includegraphics{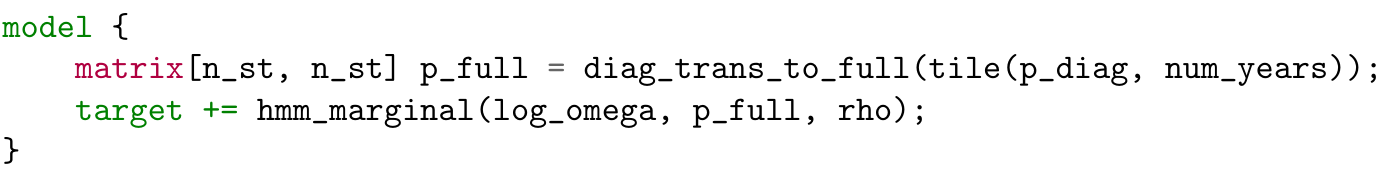}

where \texttt{target} corresponds to the log posterior.

As we're interested in the posterior probabilities (for analysis and to initialize the inital state probabilities in an iterative manner if this model is used on sequential batches of data) and sample paths, we generate the following quantities,

\includegraphics{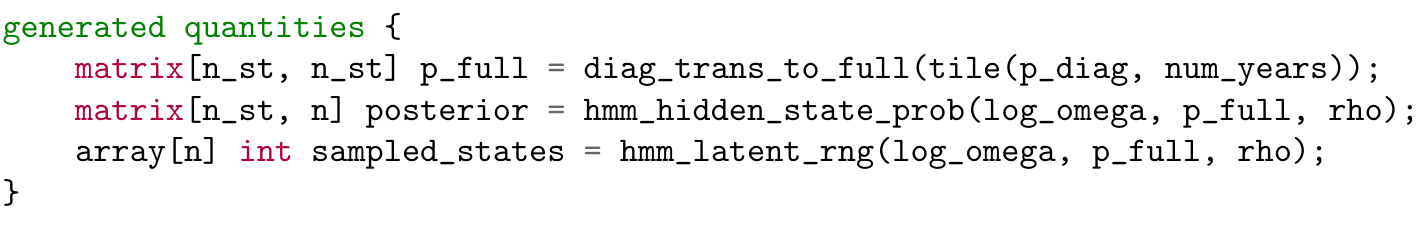}

for post-hoc analysis.

As the forward algorithm here, i.e. the function \texttt{hmm\_marginal} is inefficient for our use case, we can define a more efficient function for bidiagonal transition matrices as follows. A discussion on the computation of the forward algorithm can be found in \cite{barber, pml}.

\includegraphics{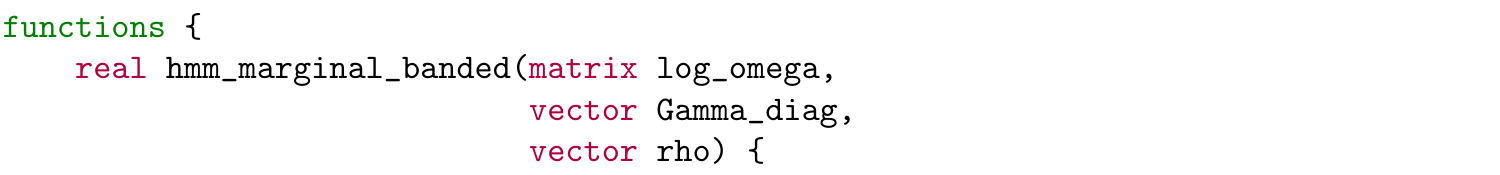}

\includegraphics[scale=0.86]{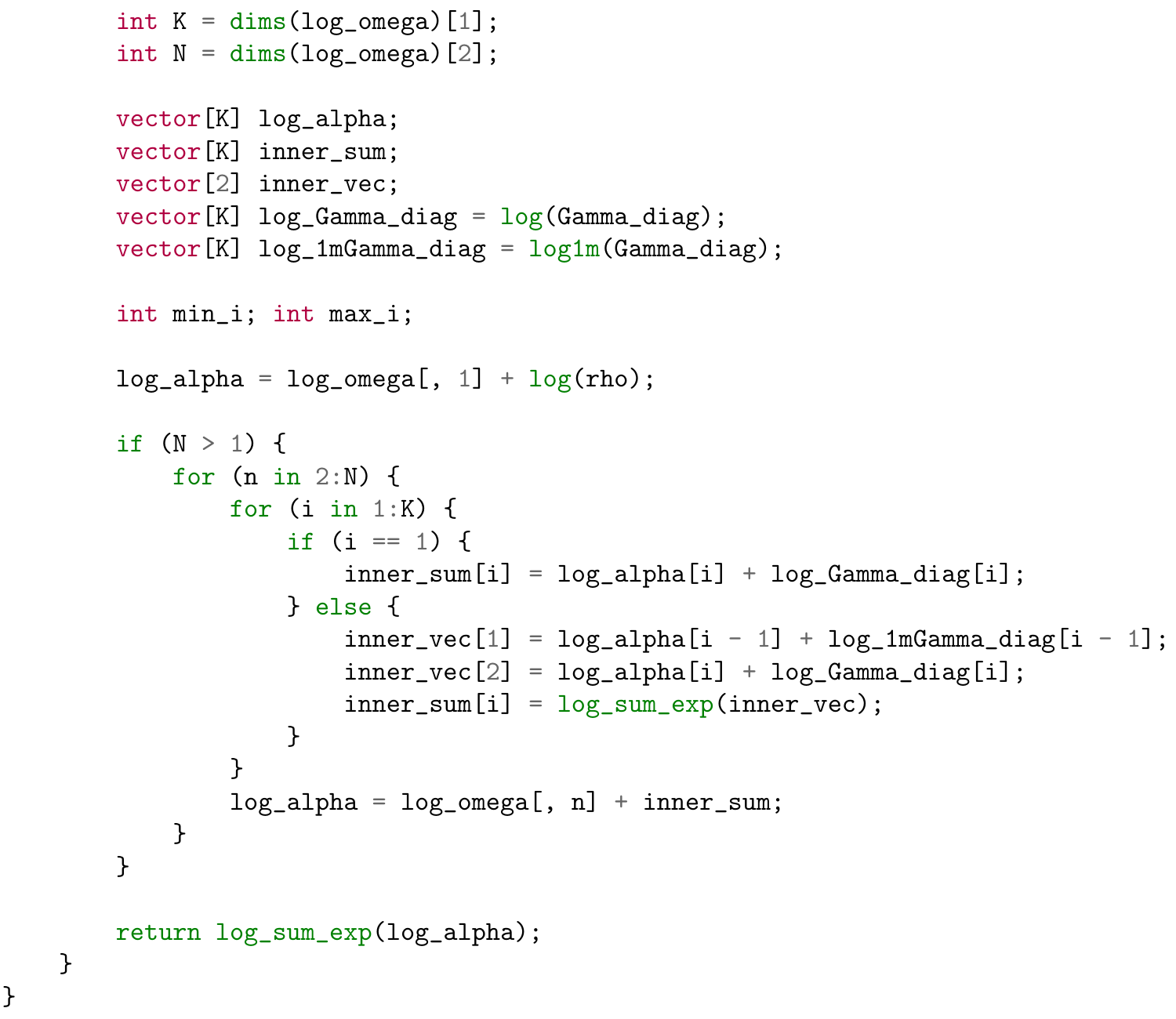}

Then, the model changes to,

\includegraphics[scale=0.86]{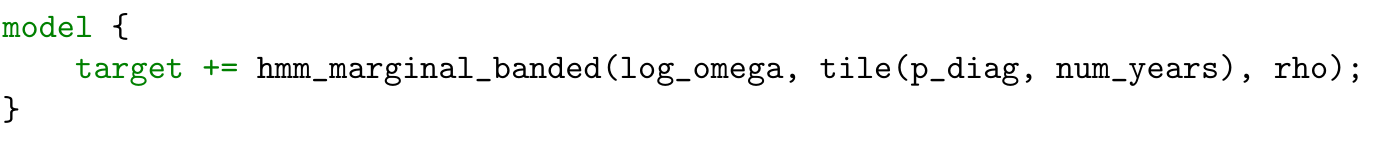}

Other convenience functions we define are as follows,

\includegraphics[scale=0.86]{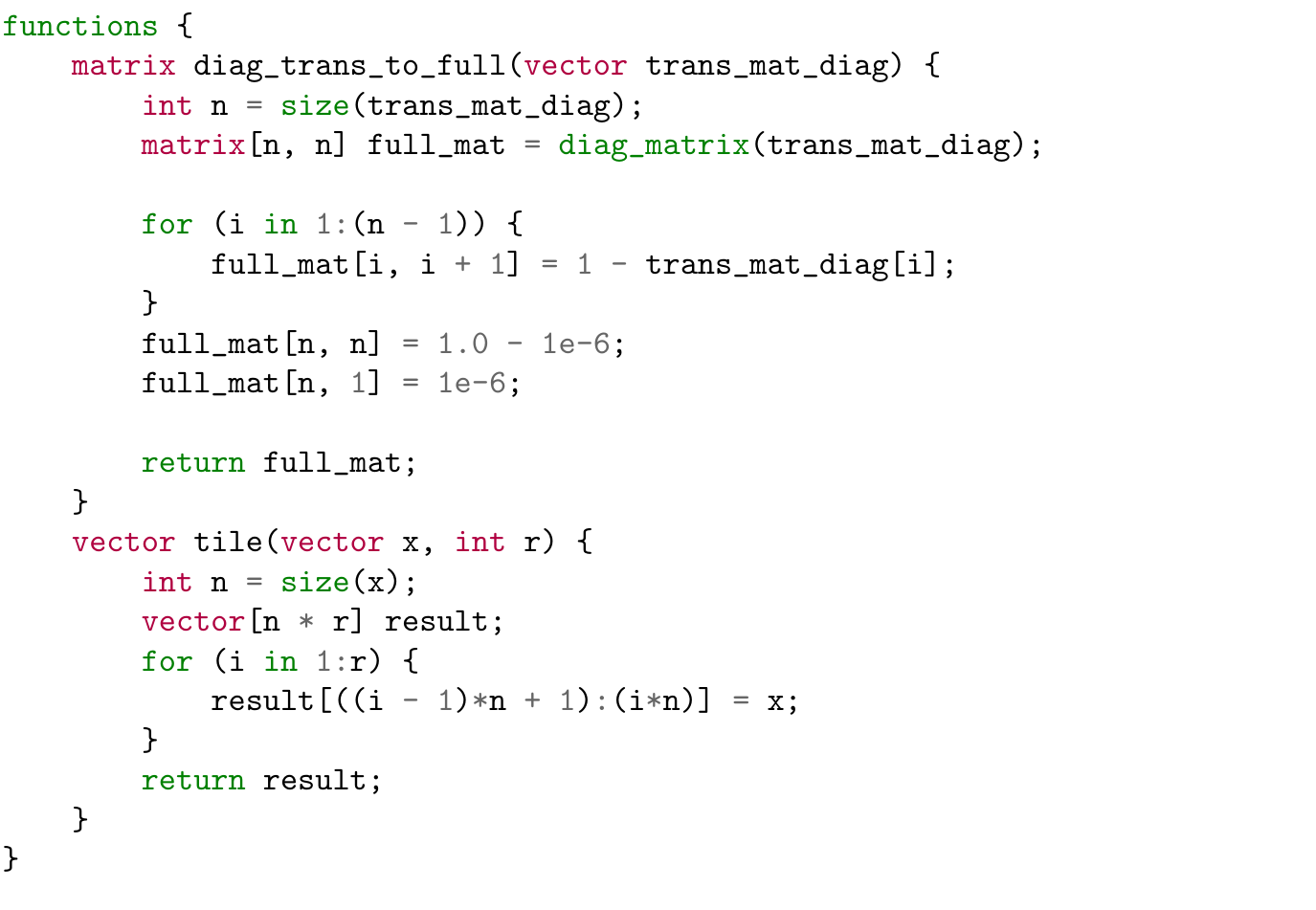}

\section{Appendix: discrete-index HMMs in Stan, hierarchical case}
\label{app:stan-hier-hmm}

To change the model to allow for changing parameters as described in \ref{subsec:tie-points}, we simply change the following aspects of the Stan model.

The parameters are extended such that there is one for every data point, with further parameters corresponding to hierarchical distribution parameters.

\includegraphics[scale=0.9]{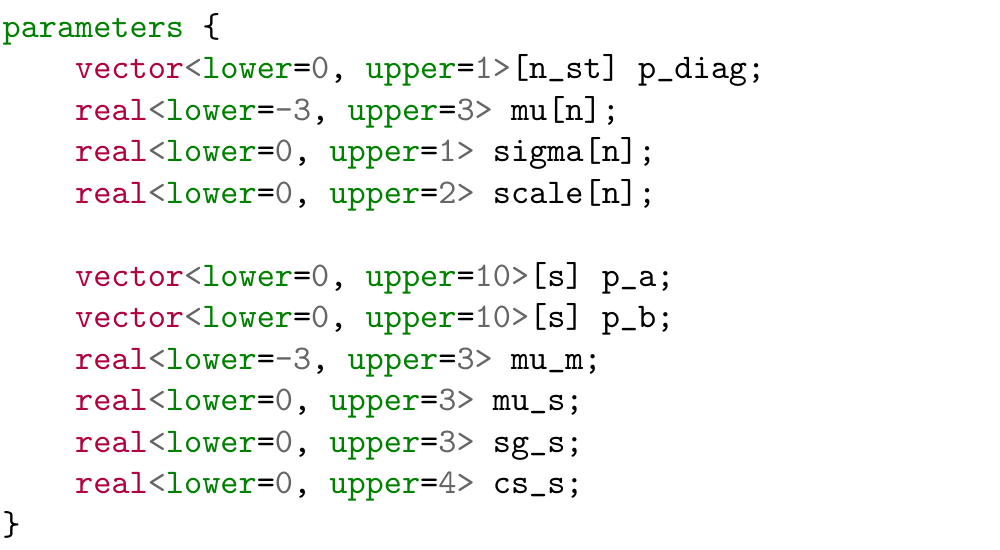}

The observation model in the transformed parameters section changes to account for the varying parameters,

\includegraphics{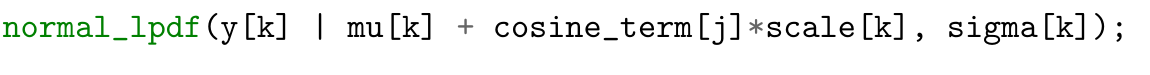}

The model block changes to account for the hierarchical priors,

\includegraphics[scale=0.9]{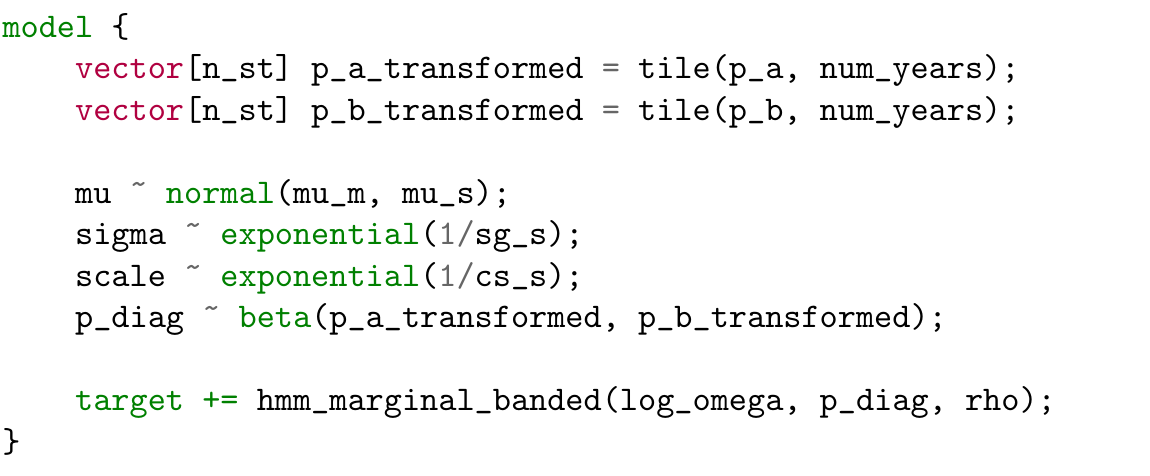}

We do not change the generated quantities block (and do not code up a more efficient version of the forward backward algorithm as the generated quantities block is only run once at the end of the inference process unlike the forward algorithm, which would be run multiple times during the inference process as the parameters are changed).

We change the inference method to variational inference instead of maximum likelihood due to the need to integrate out the hierarchical parameters during inference (as a maximum likelihood estimate of parameters such as $a_i, b_i$ in such an overparameterised model is perhaps unlikely to lie in the posterior's typical set). This is done by replacing \texttt{model.optimize} with \texttt{model.variational}. We also increased (doubled with respect to Stan's defaults) the number of gradient samples for the ELBO calculation, which was needed to obtain reasonable outputs.



\section{Appendix: cts-HMM forward algorithms in Stan}
\label{app:cts-hmm-forward}

A computationally inefficient example allowing for time varying transition matrices in the forward algorithm is shown below. In this example, the time varying transition matrix is created using \cref{eqn:cts-trans}.

\includegraphics[scale=0.9]{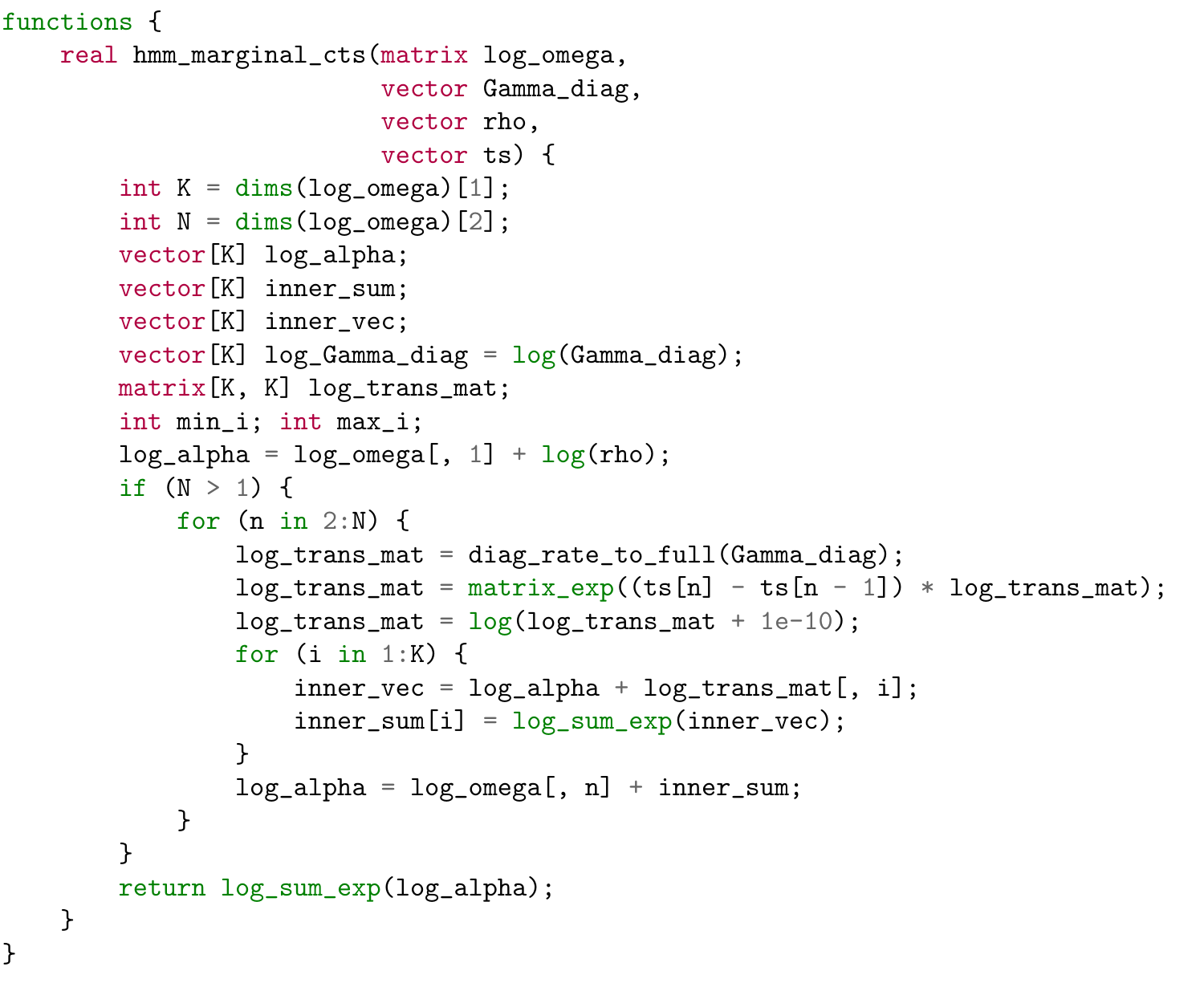}

\section{Appendix: other models}
\label{app:other-models}

We implemented the continuous time Markov chain model in \texttt{tensorflow-probability} due to out of the box support for time varying transition matrices.
We implemented our SDE models using \texttt{torchsde} due to support for differentiable SDE solvers.

The basic algorithm followed in both cases (as VI and maximum likelihood estimation are both optimization problems) is,
\begin{algorithmic}
\State $\texttt{log\_posterior\_or\_lower\_bound(params)} \gets \texttt{function(params, data) ...}$
\State $\texttt{parameters} \gets ...$ \Comment{variational params or params on which inference is done}
\State $\texttt{optimizer} \gets \texttt{Optimizer(parameters, lr)}$
\While{$\texttt{loss not converged}$} 
    \State $\texttt{loss} \gets -\texttt{log\_posterior\_or\_lower\_bound(parameters)}$
    \State $\texttt{grad} \gets \texttt{loss.grad()}$
    \State $\texttt{params} \gets \texttt{optimiser.step(params, grad)}$
\EndWhile
\\ \texttt{posthoc analysis}
\end{algorithmic}

\section{Appendix: Results of SDE models}
\label{app:sde-results}

\begin{figure}[h]
    \centering
    \includegraphics[width=0.75\textwidth]{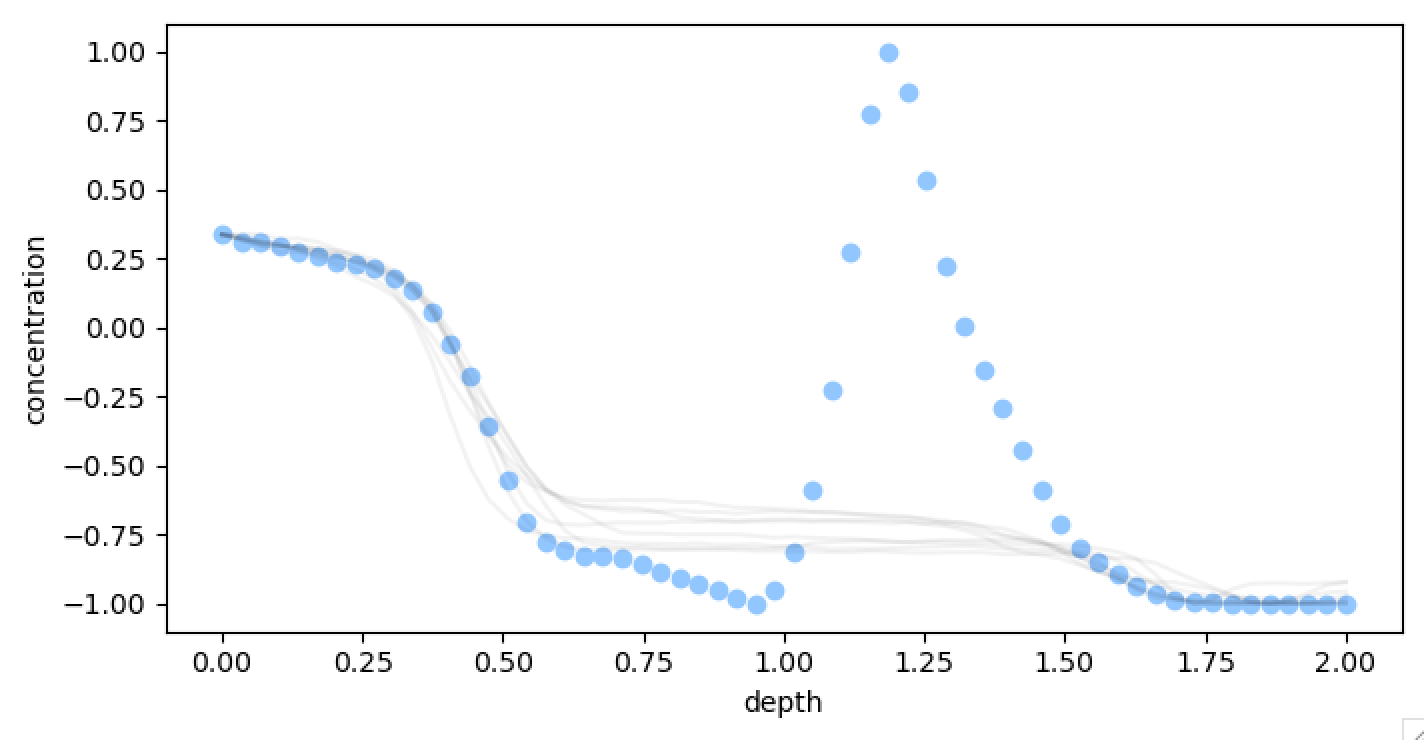}
    \caption{
        Results of our latent SDE models on synthetic data, showing a poor local maximum in the ELBO reached by VI.
        \label{fig:sde-nn}
    }
\end{figure}

Our SDE models ran into severe difficulties with inference. As an example, we assume the prior,
\begin{align*}
    \begin{bmatrix}
        dz^a_\delta \\
        dz^b_\delta \\
        dt_\delta
    \end{bmatrix}
    &=
    \begin{bmatrix}
        z^b_\delta \\
        -\lambda^2 z^a_\delta -2\lambda z^b_\delta \\
        \alpha * \sigma^+(z^a_\delta)
    \end{bmatrix} d\delta +
    \begin{bmatrix}
        0 \\
        1 \\
        10^{-2}
    \end{bmatrix} \odot d\mathbf{W}_\delta,
\end{align*}
where $\alpha$ and $\lambda$ are set to be constant, and $\sigma^+$ corresponds to the softplus operation. Note that the prior over $\mathbf{z}$ is a Matérn-3/2 Gaussian process. We use the following observation model,
$$ s_{\delta_i} | t_{\delta_i} \sim \text{Laplace}(\sin(\pi t_{\delta_i}), 0.05). $$
We tried to fit a variational posterior over $t_{\delta_i}$, using the variational SDE,
\begin{align}
\label{eqn:sde-nn}
    \begin{bmatrix}
        dz^a_\delta \\
        dz^b_\delta \\
        dt_\delta
    \end{bmatrix}
    &=
    \begin{bmatrix}
        f^a_{nn}(\mathbf{z}_\delta; t_\delta) \\
        f^b_{nn}(\mathbf{z}_\delta; t_\delta) \\
        \alpha * \sigma^+(f^c_{nn}(\mathbf{z}_\delta; t_\delta))
    \end{bmatrix} d\delta +
    \begin{bmatrix}
        0 \\
        1 \\
        10^{-2}
    \end{bmatrix} \odot d\mathbf{W}_\delta,
\end{align}
where $\mathbf{f}_{nn}$ was parameterized using a neural network, to some data simulated from the prior. The expected mean of the observation model for a few samples from the variational posterior are shown below in \cref{fig:sde-nn}.

\begin{figure}[h]
    \centering
    \includegraphics[width=0.75\textwidth]{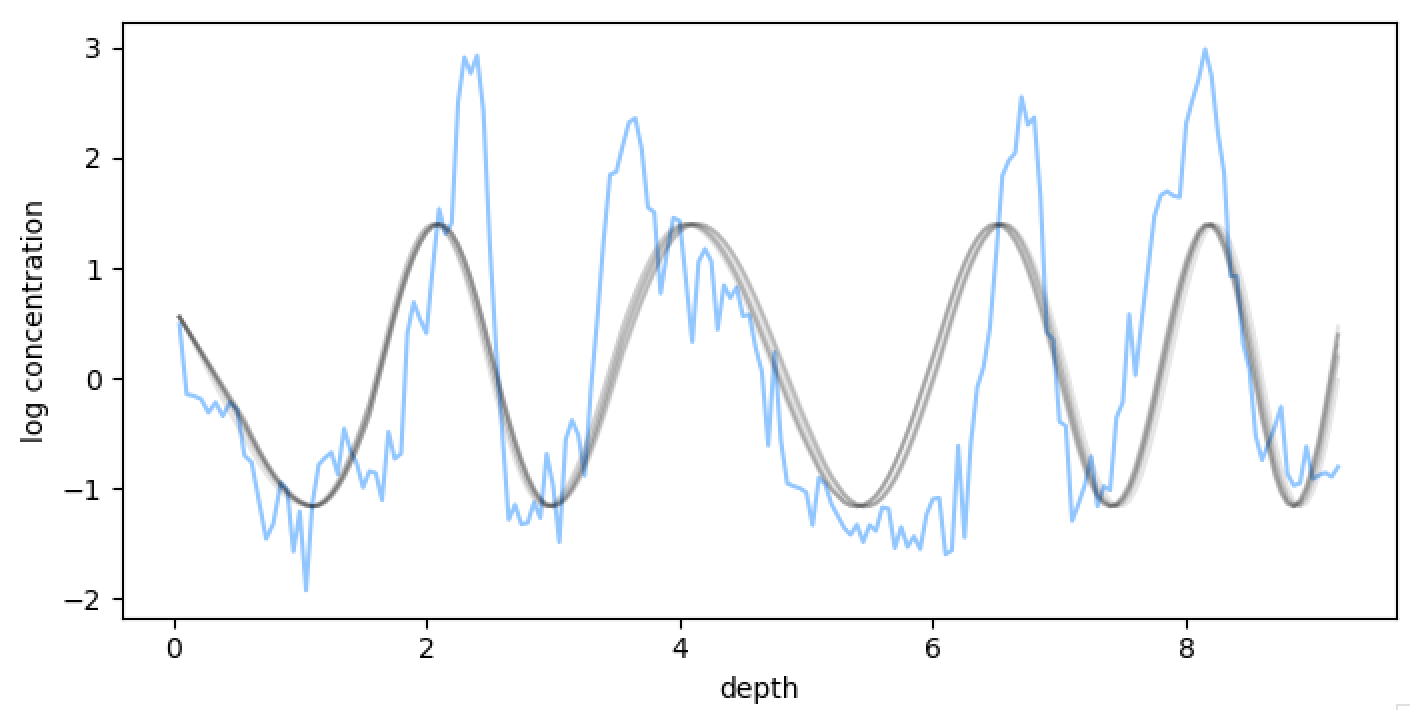}
    \caption{
        Results of our latent SDE model using spectra to initialize the prior, showing a relatively good fit to the ice core data. The blue line shows ground truth data, while the grey lines show the mean of the observation model using different samples of the posterior over time.
        \label{fig:tf-init}
    }
\end{figure}

We also tried to use sparse Gaussian processes for placing priors and variational approximations on $\mathbf{z}_\delta$, utilizing random Fourier features \citep{rff} to sample functions from these GPs to work seamlessly with torchsde, finding no improvement in results. Pre-fitting a neural-network based SDE prior (using spectra derived from the data) however results in a better fit to the ice core data (illustrated below in \cref{fig:tf-init}), however, results are still very (impractically) sensitive to initialisation.

Future work can also involve exploration of Kalman filtering algorithms for this problem, as our SDE priors are valid models that may be assumed for state estimation in extended Kalman filtering.

\end{document}